\documentclass[10pt,twocolumn,letterpaper]{article}

\usepackage[pagenumbers]{cvpr}              

\usepackage{graphicx}
\usepackage{amsmath}
\usepackage{amssymb}
\usepackage{booktabs}
\usepackage[table]{xcolor}

\usepackage[pagebackref,breaklinks,colorlinks]{hyperref}

\usepackage[capitalize]{cleveref}
\crefname{section}{Sec.}{Secs.}
\Crefname{section}{Section}{Sections}
\Crefname{table}{Table}{Tables}
\crefname{table}{Tab.}{Tabs.}

\usepackage{array,multirow,graphicx}

\def\cvprPaperID{1} 
\def\confName{CVPR}
\def\confYear{2022}

\newcommand{\mysection}[1]{\vspace{2pt}\noindent\textbf{#1}}

\newcommand\blfootnote[1]{%
  \begingroup
  \renewcommand\thefootnote{}\footnote{#1}%
  \addtocounter{footnote}{-1}%
  \endgroup
}

\begin{document}

\title{SoccerNet-Caption: Dense Video Captioning\\for Soccer Broadcasts Commentaries}

\author{Hassan Mkhallati$^{*,1}$
\quad
Anthony Cioppa$^{*,2,3}$  
\quad
Silvio Giancola$^{*,3}$  
\and
Bernard Ghanem$^{3}$  
\quad
Marc Van Droogenbroeck$^{2}$  
\and$^1$   {\small Université Libre de Bruxelles}
\quad $^2$ {\small University of Liège}
\quad $^3$ {\small KAUST}
}
\maketitle

\begin{abstract}
Soccer is more than just a game - it is a passion that transcends borders and unites people worldwide. From the roar of the crowds to the excitement of the commentators, every moment of a soccer match is a thrill. 
Yet, with so many games happening simultaneously, fans cannot watch them all live. Notifications for main actions can help, but lack the engagement of live commentary, leaving fans feeling disconnected.
To fulfill this need, we propose in this paper a novel task of dense video captioning focusing on the generation of textual commentaries anchored with single timestamps.
To support this task, we additionally present a challenging dataset consisting of almost 37k timestamped commentaries across 715.9 hours of soccer broadcast videos. 
Additionally, we propose a first benchmark and baseline for this task, highlighting the difficulty of temporally anchoring commentaries yet showing the capacity to generate meaningful commentaries.
By providing broadcasters with a tool to summarize the content of their video with the same level of engagement as a live game, our method could help satisfy the needs of the numerous fans who follow their team but cannot necessarily watch the live game.
We believe our method has the potential to enhance the accessibility and understanding of soccer content for a wider audience, bringing the excitement of the game to more people.

\blfootnote{\hspace{-5mm}\textbf{(*)} Equal contributions. Data/code available at \href{https://www.soccer-net.org}{www.soccer-net.org}.
\\Contacts: $\{$name.surname$\}$@kaust.edu.sa / soccernet@uliege.be.
}

\end{abstract}

\begin{figure}[t]
    \centering
    \includegraphics[width=\linewidth]{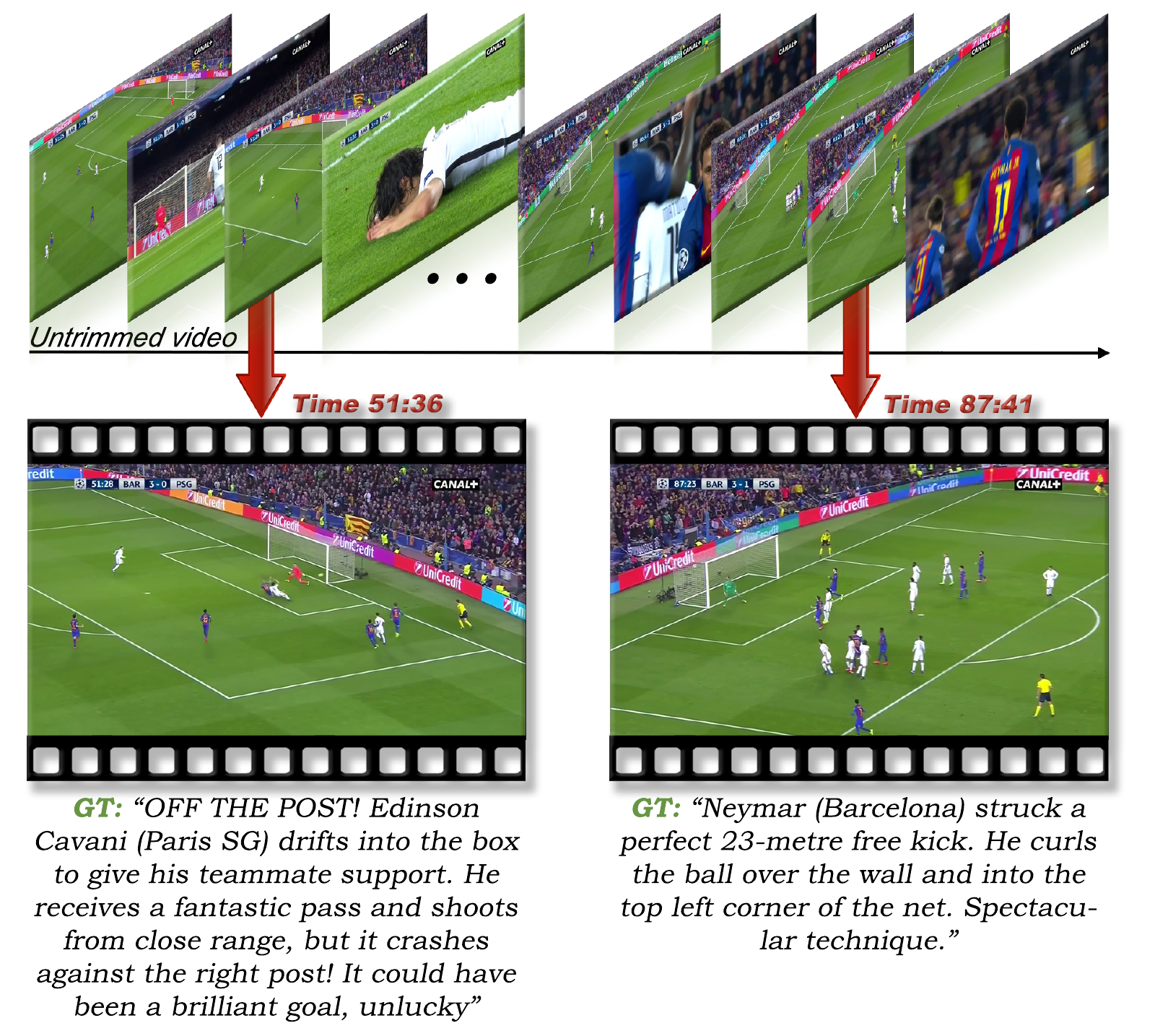}
    \caption{
    \textbf{SoccerNet-Caption.}
    We provide a large-scale dataset for Single-anchored Dense Video Captioning (SDVC) in untrimmed soccer broadcast videos. 
    Our SoccerNet-Caption dataset is composed of $36{,}894$ textual commentaries, temporally anchored within $715.9$ hours of soccer broadcasts. The comments describe the events occurring in the soccer game with rich factual, emotional, and sensational content.
    }
    \label{fig:graphical_abstract}
\end{figure}

\section{Introduction}
\label{sec:introduction}

Over the past decade,  the quantity and quality of sports data
has increased rapidly. 
This explosion has been driven by the benefits of automated analysis in various applications such as player performance, providing insights into game strategy~\cite{Tuyls2021Game}, and audience involvement. 
Live text commentaries provide a rich summary of the game to increase fan engagement for those who do not have time or the possibility to watch the game.
However, they are usually exclusive to major professional leagues, while other games are often left out.
Moving towards automated solutions relying on already available equipment is therefore essential for lower leagues and amateur soccer.
%
Recently, there has been a growing interest in the research community to automatically generate text based on videos.
This task, called Dense Video Captioning (DVC), represents a significant research challenge due to the memory footprint of video data and the complexity of natural language. 
Besides, most research focuses on generic descriptions of events and activities in open-world scenes.
However, in soccer, commentaries need to include rich factual, emotional, and even sensational content to engage the fan.
Sports is therefore the perfect playground for research in the video and language domain.

In this paper, we publicly release \textit{SoccerNet-Caption}, the first dataset for dense video captioning in soccer broadcast videos. In particular, we provide $36{,}894$ temporally-anchored rich textual commentaries describing $715.9$ hours of soccer games. Some examples of comments from our dataset are shown in Figure~\ref{fig:graphical_abstract}.
Along with the data, we introduce the new task of Single-anchored Dense Video Captioning (SDVC), consisting in generating localized captions describing the soccer game. 
As a first benchmark, we propose a two-stage approach including an action spotting module and a captioning module. 
More specifically, the spotting module produces temporal proposals for the captions.
Then, the videos are trimmed around the proposals and passed to the captioning module to generate captions.
We show in Figure~\ref{fig:qualitative} that our approach allows to generate relevant captions to describe the game with rich semantics, but that the challenge is still open for major improvements.

\mysection{Contributions.} We summarize our contributions as follows:
\textbf{(i)} We publicly release the largest dataset of soccer videos annotated with timestamped textual commentaries describing the game. 
\textbf{(ii)} We define the novel task of Single-anchored Dense Video Captioning (SDVC), where captions are anchored with a single timestamp and need to be generated in long untrimmed videos. 
\textbf{(iii)} We propose a first benchmark to tackle this task and provide a thorough ablation study and analysis.

\section{Related Works}
\label{sec:relatedworks}

\mysection{Sport understanding.}
The challenging aspect of sports video understanding has contributed to its growing popularity as a research focus~\cite{Moeslund2014Computer, Thomas2017Computer}. At first, methods focused on video classification~\cite{Wu2022ASurvey-arxiv}, including the recognition of specific actions~\cite{Saraogi2016Event,Khan2018Learning}, or segmentation of different game phases during the game~\cite{Cioppa2018ABottomUp}. 
More recently, the task of action spotting was introduced by Giancola~\etal~\cite{Giancola2018SoccerNet}, aiming at providing the precise localization of specific actions within an untrimmed soccer broadcast video. Several methods were proposed to automate this process, for instance, using a context-aware loss ~\cite{Cioppa2020AContextAware}, camera calibration and player localization~\cite{Cioppa2021Camera}, end-to-end training~\cite{Hong2022Spotting-arxiv}, spatio-temporal encoders~\cite{Darwish2022STE}, graphs-based methods~\cite{Cartas2022AGraphbased}, transformer-based methods~\cite{Zhu2022ATransformerbased}, or anchor-based methods~\cite{Soares2022Action-arxiv, Soares2022Temporally}.  
Other methods focused on other aspects of sports understanding such as player detection~\cite{Vandeghen2022SemiSupervised}, player tracking~\cite{Maglo2022Efficient} and identification~\cite{Vats2022IceHockey, Somers2023Body}, tactics analysis in soccer and fencing~\cite{Suzuki2019Team, Zhu2022FenceNet}, pass feasibility~\cite{ArbuesSanguesa2020Using}, 3D ball localization for basketball~\cite{VanZandycke20223DBall}, or 3D shuttle trajectory reconstruction for badminton videos~\cite{Liu2022MonoTrack}.

To support this research, large-scale datasets have been released, including the ones of Pappalardo et al.~\cite{Pappalardo2019}, Yu et al.~\cite{Yu2018Comprehensive}, SoccerTrack~\cite{Scott2022SoccerTrack}, SoccerDB~\cite{Jiang2020SoccerDB}, and DeepSportRadar~\cite{VanZandycke2022DeepSportradarv1}.
The SoccerNet dataset, introduced by Giancola~\etal~\cite{Giancola2018SoccerNet}, includes benchmarks for $10$ different tasks related to soccer understanding, such as action spotting~\cite{Deliege2021SoccerNetv2}, camera calibration~\cite{Cioppa2022Scaling}, and player re-identification~\cite{Cioppa2022Scaling}. Cioppa~\etal also introduce the task of player tracking in long sequences, including long-term re-identification~\cite{Cioppa2022SoccerNetTracking}. Yearly competitions are organized on this dataset to promote research sports~\cite{Giancola2022SoccerNet}. Our novel SDVC task data is part of the 2023 challenges.


\mysection{Video-Language Datasets.} 
Initially, research on video combined with language focused on video tagging, including actions and objects~\cite{Rafiq2021Video, Damen2021Rescaling, Regneri2013Grounding}.
With the successes in image captioning~\cite{Chen2020UNITER, pmlr-v139-radford21a} (\ie describing an image with natural language), research has shifted towards deep learning approaches for video captioning.
%
Large-scale multimodal datasets have been introduced thanks to the rise of automatic speech recognition techniques and video-sharing platforms. 
Youtube has been a major data source for \textbf{YouTube-8M}~\cite{AbuElHaija2016YouTube8M-arxiv}, 
\textbf{HowTo100M}~\cite{Miech2019HowTo100M}, and 
\textbf{ViTT}~\cite{Huang2020Multimodal-arxiv}. 
Other datasets focused on domain-specific videos such as cooking: \textbf{Youcook2}~\cite{Zhou2018Towards} and \textbf{Tacos}~\cite{Das2013AThousand}, or movies:
\textbf{MAD}~\cite{Soldan2022MAD}. 
Further efforts have been proposed for egocentric vision such as \textbf{EPIC-KITCHENS-100}~\cite{Damen2021Rescaling}, and \textbf{Ego4D}~\cite{Grauman2022Ego4D}.
On top of those datasets, various tasks have emerged such as video-language grounding~\cite{Hendricks2017Localizing, Bain2021Frozen, Soldan2022MAD}, video question answering~\cite{Tapaswi2016MovieQA, Grauman2022Ego4D}, video clip captioning~\cite{Rohrbach2015ADataset, Yu2018FineGrained, Gao2017Video, Hammoudeh2022Soccer} and dense video captioning~\cite{Krishna2017DenseCaptioning-arxiv, Zhou2018Towards}.

\mysection{Dense Video Captioning.} 
Krishna~\etal~\cite{Krishna2017DenseCaptioning-arxiv} introduced this task, which consists in captioning temporally localized (start and end frame) activities in untrimmed video. 
This task differs from traditional video captioning~\cite{Gao2017Video}, where a single caption is generated for short videos, and dense image captioning~\cite{Johnson2016DenseCap}, where captions describe different regions of an image. 
Currently, \textbf{YouCook2}~\cite{Zhou2018Towards}, including recipes for non-overlapping sequential events, and \textbf{ActivityNet-Captions}~\cite{Krishna2017DenseCaptioning-arxiv}, including open-domain overlapping activities, are the standard benchmarks for dense video captioning.
Our dataset introduces a new task of single-anchored dense video captioning for soccer video comment generation, which requires richer factual, emotional, and sensational comments.
Traditionally, the solutions proposed for dense video captioning involve a two-stage  ``detect-then-describe'' framework.
The first module produces temporal proposals, and the second module generates captions around these proposals~\cite{Zhou2018Towards, Krishna2017DenseCaptioning-arxiv, Iashin2020Multimodal-arxiv, Iashin2020ABetter-arxiv}.
With the rise of large datasets, many efforts have focused on efficient pre-training of models that combine both language and video~\cite{Bain2021Frozen, Miech2020EndtoEnd, Yang2021Just, alayrac2022flamingo, Seo2022Endtoend, Zhang2022Unifying-arxiv, Huang2020Multimodal-arxiv}. 
Recent works also tried ``YOLO-like''~\cite{Redmon2016YouOnly} approaches where localization and captioning are generated in one shot~\cite{Yang2023Vid2Seq-arxiv, Wang2021EndtoEnd}.
Here, we propose a two-stage approach based on a pre-trained video encoder.



\section{Dataset}
\label{sec:dataset}

\mysection{Data collection.}
Our SoccerNet-Caption dataset comprises $471$ SoccerNet untrimmed broadcast games, including the top five European leagues (EPL, La Liga, Ligue 1, Bundesliga, and Serie A) as well as the Champions League from 2014 to 2017.
All videos are available at $25$fps in two resolutions: $224$p and $720$p, alongside frame features at $2$fps pre-extracted using \eg ResNET or the Baidu feature encoder~\cite{Zhou2021Feature} at $1$fps, following SoccerNet's original data format.
Since the role of the producer is to select the right camera to convey the story of the game to the viewer in the best possible way, these  broadcasted games are perfectly suited for a commentary generation task.

In this work, we provide novel textual comments embedded in time describing the game.
We collect those comments by scrapping the flashscore website for $471$ out of the $500$ games included in the SoccerNet dataset. The commentaries for the remaining games were not unavailable.
The comments typically describe in a few sentences the main events occurring at specific times in the game, by giving insights about the involved players or teams, and the sequence of actions that led to this situation in a rich factual, emotional, and sensational way.
Our data collection efforts resulted in $36{,}894$ timestamped comments across $715.9$ hours of video footage, including some metadata such as the type of event the comment relates to (\eg an action, a fun fact, \etc).

Alongside these textual comments, we also collected metadata about the game, including the list of all players with their jerseys number, the referees' name, and the teams' name. These metadata also includes the starting lineups for each team, with their tactics, the $11$ starting players, the substitutes, and any events associated with a player, such as goals, assists, substitutions, and yellow/red cards.

\mysection{Data anonymization.}
Following the traditional captioning dataset, we provide an anonymized version of the captions, where each player, referee, team, and coach names are replaced with generic tokens. 
In fact, most captioning methods are not suited to recognize the exact identity of the people shown in the videos.
Hence, without including specific modules for identity classification, player tracking, and re-identification, it would be almost impossible to generate the correct names.
However, we still provide the original captions for future research.

To anonymize the dataset, we leverage the game metadata to retrieve the team, coach, referee, and player names. 
Then, we automatically search through the comments for these specific names and replace them with a generic token ([TEAM], [COACH], [REFEREE], or [PLAYER]).
Sometimes, the player and coach names available in the lineups may differ from those mentioned in the comments as some particles or longer names may be truncated. 
To retrieve those names, we used advanced string-matching techniques to identify and reconcile any discrepancies between the names in the lineups and those mentioned in the comments. 
Finally, for the remaining names (\eg compound surnames are formulated differently between the lineups and the comments), we manually refined the annotation.
This approach ensures that we accurately link each name to the correct token.

\begin{figure}
    \centering
    \includegraphics[width=\linewidth]{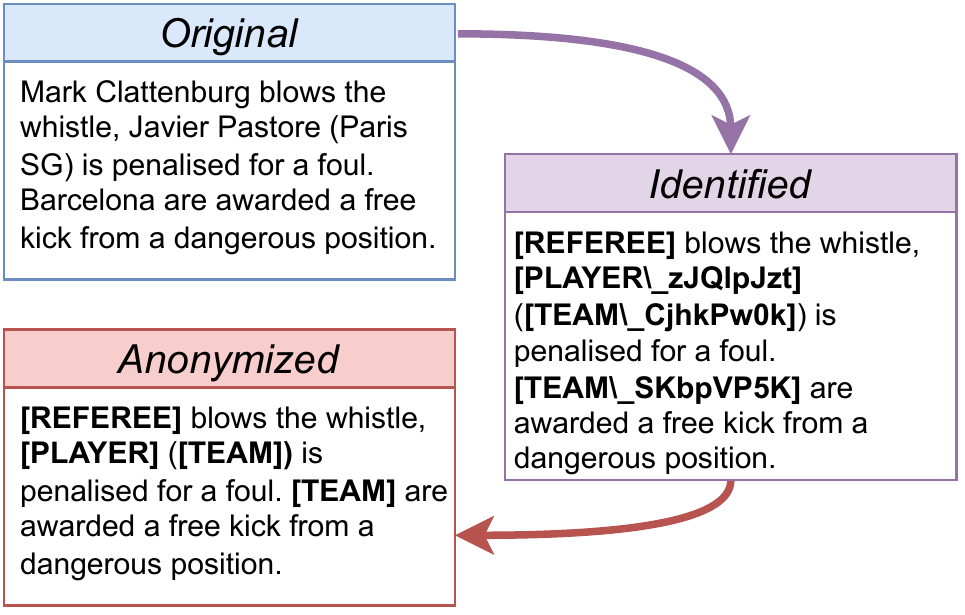}
    \caption{\textbf{Comment anonymization.} We provide three versions for each comment. The original commentary, an identified version where each player is associated with a unique id token, and an anonymized version where each entity is replaced by a specific token: [TEAM], [COACH], [REFEREE], and [PLAYER].}
    \label{fig:comment examples}
\end{figure}

Eventually, we provide an intermediate anonymization, where each player is identified with a unique id (uid) inside the token ([Player\_uid]). This alleviates the issue of the different names for the same player in the original captions and maintains his identity.
%
Examples of such comments are provided in Figure~\ref{fig:comment examples}.

\mysection{Data format.}
Following the SoccerNet format, we organize our textual annotations into individual JSON files for each game.
Each file contains a dictionary that includes all metadata of the game (\eg the starting lineups for each team, with their tactics, the $11$ starting players, the substitutes, their jersey numbers, \etc) and the list of annotated comments. Each annotation is associated with a timestamp, the three versions of the comment (original description, identified, and anonymized), a boolean value indicating whether the comment is related to a key moment of the game, and a contextual label
(\eg corner, substitution, yellow card, whistle, soccer ball, time, injury, fun fact, attendance, penalty, red card, own goal, or missed penalty).

\begin{figure}[t]
    \includegraphics[width=\linewidth]{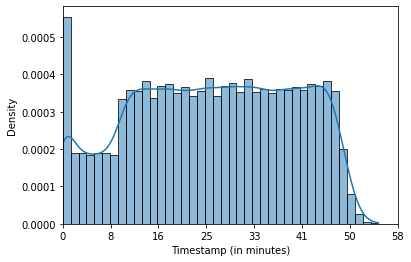}
    \caption{\textbf{Distribution of the comments.} Most comments are uniformly scattered in each half-time, except at the start of the game where a peak is followed by fewer comments for $10$ minutes.}
    \label{fig:distributionpergame}
\end{figure}

\mysection{Data statistics.}
SoccerNet-Caption dataset contains an average of $78.33$ temporally localized comments per game, resulting in a total of $36{,}894$ captions for the entire dataset. This is equivalent to almost one comment every minute. 
%
As can be seen in Figure~\ref{fig:distributionpergame}, the distribution of the comments within a single game over time shows a peak at the start of the game that usually corresponds to the comment related to the first whistle of the referee. Then, there is a period of fewer comments in the first $10$ minutes compared to the rest of the half-time that follows a uniform distribution. This shows that no bias can be used to find a good location for the comments, except at the very start of the game.

\begin{figure}[t]
    \centering
    \includegraphics[width=0.95\linewidth]{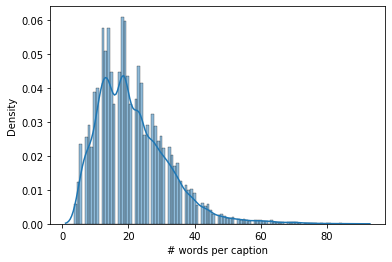}
    \caption{\textbf{Distribution of the number of words per comment} This plot shows that the number of words per comment follows a long tail distribution with $21.38$ words on average.}
    \label{fig:wordspercomment}
\end{figure}

\begin{figure}[t]
    \includegraphics[width=\linewidth]{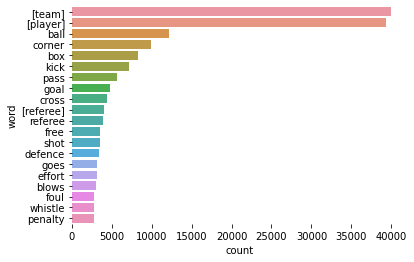}
    \caption{\textbf{Distribution of the most common words.} The most frequent words are the names of the teams and the players, followed by words semantically related to soccer verbs and soccer elements. There is a high imbalance in the distribution.}
    \label{fig:worddistribution}
\end{figure}

Finally, we analyze the content of each comment on a textual and semantic level. Figure~\ref{fig:wordspercomment} shows that the number of words per comment ranges from $4$ to $93$ words following a long tail distribution with $21.38$ words on average. As mentioned, soccer has its own specific terminology for describing events. We can observe in Figure~\ref{fig:worddistribution} that, apart from the generic tokens, the most commonly used words are soccer action-related verbs (\eg kick, pass, cross) or soccer-related nouns (\eg corner, box, goal). 
Additionally, it is important to note the over-representation of player and team names in the comments which motivates the use of anonymized comments.


\mysection{Novelty.}
We compare our dataset with other recent captioning and dense video captioning datasets in Table~\ref{tab:comparisonwithdatasets}.
Our dataset provides the longest videos on average by a large margin. Processing such long videos in an end-to-end approach is still an open challenge in most video understanding tasks, which makes our dataset the perfect playground to innovate. SoccerNet-Caption also ranks third in terms of total video length, making it a large-scale dataset for video and language training. The other soccer-related dataset either focus on captioning short clips or highlights. 
SoccerNet-Caption is the first dataset to anchor the comments with a single timestamp instead of a bounding box with a start and end timestamp for each comment. Following the work of SoccerNet on action spotting, we believe that it is arduous to annotate when a specific action starts and ends in soccer. 
 %
Finally, compared to more generic datasets, soccer commentaries require richer content, including emotional or sensational sentences. By providing the original comments with the game metadata, we aim to push research toward identifying the players in the video to accurately describe what is happening in the game. 

\begin{table*}[t]
\renewcommand*{\arraystretch}{0.85}
 \centering
 \resizebox{\linewidth}{!}{
 \begin{tabular}{l||c|c|c|c|c|c|c}
 \rowcolor[HTML]{EFEFEF}
Name  & Domain & \# Video &  Avg Duration (s) &  Total Duration (hr)    & \# Sentences & Anchors & Task \\ \midrule
ActivityNet-Caption \cite{Krishna2017DenseCaptioning-arxiv}  & Open & 20k  & 180  & 849   & 100k  & $[t_s, t_e]$ & DVC \\ \rowcolor[HTML]{EFEFEF}
Youcook2 \cite{Zhou2018Towards}                         & Cooking                       & 2k   & 30   & 176   & 15.4k & $[t_s, t_e]$ & DVC\\ 
TACoS \cite{Das2013AThousand} / TACoS-Multilevel \cite{Rohrbach2014Coherent} & Cooking  & 127/185  & 360  & 15.9/27.1  & 18.2k/52.5k & $[t_s, t_e]$ & Retrievial \\ \rowcolor[HTML]{EFEFEF}
Charades-STA \cite{Sigurdsson2016Hollywood}             & Human                         & 9.8k & 30   & 82.01 & 27.7k & $[t_s, t_e]$ & Retrievial \\
VideoStory \cite{Gella2018ADataset}                     & Social Media                  & 20k & 70 & 396 & 123k & $[t_s, t_e]$ & Storytelling\\ \rowcolor[HTML]{EFEFEF}
ViTT \cite{Huang2020Multimodal-arxiv}                   & Cooking + open                & 8.2k & --    & --     & -- (Tag)     & $[t_s, t_e]$ & DVC \\ 
Epic Kitchen-100 \cite{Damen2021Rescaling}              & Cooking-Ego                   & 700  & 514  & 100   & --     & $[t_s, t_e]$ & Action Recognition\\ \rowcolor[HTML]{EFEFEF}
Ego4D \cite{Grauman2022Ego4D}                           & Ego                           & 9.6k & 1369.8     & 3,670 & 3.85M & $[t_s, t_e]$ & Moment Queries \\ 
DiDeMo \cite{Hendricks2017Localizing}                   & Open-human                    & 10K  & 30   &  88.7     & 40.5k & $[t_s, t_e]$ & Retrievial \\ \rowcolor[HTML]{EFEFEF}
MAD \cite{Soldan2022MAD}                                & Movie                         & 650  & 6646.2 & 1207.3    & 384.6k & $[t_s, t_e]$ & Grounding \\
Fine-grained Sports Narrative \cite{Yu2018FineGrained}  & Basketball                    & 2K   & -- & -- & 6520  & -- & Video Captioning \\ \rowcolor[HTML]{EFEFEF}
Soccer captioning \cite{Hammoudeh2022Soccer}            & Soccer action clips           & 22k clip-action & -- & -- & 22k & -- & Video Captioning \\ 
GOAL (Qi~\etal~\cite{Qi2023GOAL-arxiv})                 & Soccer action clips           & 8.9k &  10.31 &  25.5  & 22k &  --  & KGVC\\ \rowcolor[HTML]{EFEFEF}
GOAL (Suglia~\etal~\cite{Suglia2022Going-arxiv})        & Soccer highlights             & 1.1k & 238  &  73.1   & 53k   & -- & Video Captioning\\ \midrule
SoccerNet-caption (ours)                                & Soccer games                  & 942  & 2735.9 & 715.9 & 36,894   & $t$ & SDVC 
 \end{tabular}
 }
 \caption{
 \textbf{Comparison of SoccerNet-Caption with other captioning datasets.} 
 Our dataset contains the second longest video sequences as well as the third longest total video length. This shows that SoccerNet-Caption is a great dataset for research in dense video captioning. Also, it is the first dataset on untrimmed soccer broadcast games, unlike GOAL (Suglia~\etal~\cite{Suglia2022Going-arxiv}) which only focuses on soccer highlights and GOAL (Qi~\etal~\cite{Qi2023GOAL-arxiv})  which only focuses on soccer actions for Knowledge-Grounded Video Captioning (KGVC)~\cite{Qi2023GOAL-arxiv}.}
 \label{tab:comparisonwithdatasets}
\end{table*}
\section{Single-anchored Dense Video Captioning}
\label{sec:Methodology}

\mysection{Task.}
We define the novel task of  Single-anchored Dense Video Captioning (SDVC) as follows:
Given a video, spot all instants where a comment should be anchored and generate sentences describing the events occurring around that time using natural language.
This is different from the previously defined Dense Video Captioning (DVC) task \cite{Krishna2017DenseCaptioning-arxiv}, where the captions have temporal boundaries (start and end timestamps).
For soccer games, this task is particularly challenging, as it requires describing complex sequences of actions involving subtle player movements, rather than well-separated activities with clearly defined boundaries. 







\mysection{Metric.}
Defining a metric for this task requires evaluating both the temporal accuracy of the detected anchors and the quality of the generated commentaries. 
Finding the exact timestamp for a commentary is challenging as it depends on the game evolution or certain actions that need to be highlighted, rather than being associated with a specific action like the events defined in the action spotting task.
Therefore, we need to include some tolerance around the ground-truth action spot.
Additionally, evaluating the quality of generated commentary is not trivial as the expressions used are semantically more related compared to an open vocabulary. Hence, subtle variations in the chosen words need to be accurately evaluated when describing the game.

In the literature, several metrics have been proposed to evaluate dense video captioning methods. The SODA metric~\cite{Fujita2020SODA} evaluates the video narrative by finding the temporally optimal matching between generated and reference captions. 
Hammoudeh~\etal~\cite{Hammoudeh2022Soccer} proposed another approach that focuses on the precision and recall of generated words with specific expressions defined beforehand and matched with the ground truth. 
This approach is useful in terms of semantic accuracy in captions but strongly depends on the chosen words dictionary. 
EMScore~\cite{Shi2022EMScore} focuses on the consistency between the video and candidate captions, relying on an Embedding Matching-based score. 
However, the EMScore depends on the performance of the used vision-language pre-trained (VLP) model. 
Since no VLP models have been trained on sports videos yet, we cannot use one of these metrics for our dataset.
In order to quantify the spotting quality of the generated captions, we use the mAP@$\delta$ for action spotting introduced in Giancola~\etal~\cite{Giancola2021Temporally}, where $\delta$ is the tolerance in seconds.

For the SDVC task, we first choose the metric proposed in ActivityNet-captions (that still has a consensus within the community) and adapt it for our single-anchored task as follows: for each ground-truth caption in a video, we build a time window with a chosen tolerance centered on its timestamp. 
We then use established captioning evaluation metrics (METEOR~\cite{Lavie2007Meteor}, BLEU~\cite{Papineni2001BLEU}, ROUGE~\cite{Lin2004ROUGEAP}, and CIDEr~\cite{Vedantam2015CIDEr}) to estimate the language similarity between all generated captions with any ground-truth caption for which its timestamps fall within a $\delta$ tolerance. The performances are finally averaged over the video and the dataset. We call our metric METEOR@$\delta$ (resp. BLEU@$\delta$,  ROUGE@$\delta$, and CIDEr@$\delta$).
As a second metric, we similarly adapt the SODA\_c~\cite{Fujita2020SODA} metric by adding a time window around the ground-truth and generated captions.
\section{Benchmarks}
\label{sec:experiments}

In this section, we benchmark a first baseline model on our task of Single-anchored Dense Video Captioning (SDVC). Particularly, we study the performance of several architectures and hyperparameters of our model on the spotting, captioning, and global SDVC task. We finally provide qualitative results on real sequences.

\subsection{SDVC baseline}
\begin{figure}
    \centering
    \includegraphics[width=\linewidth]{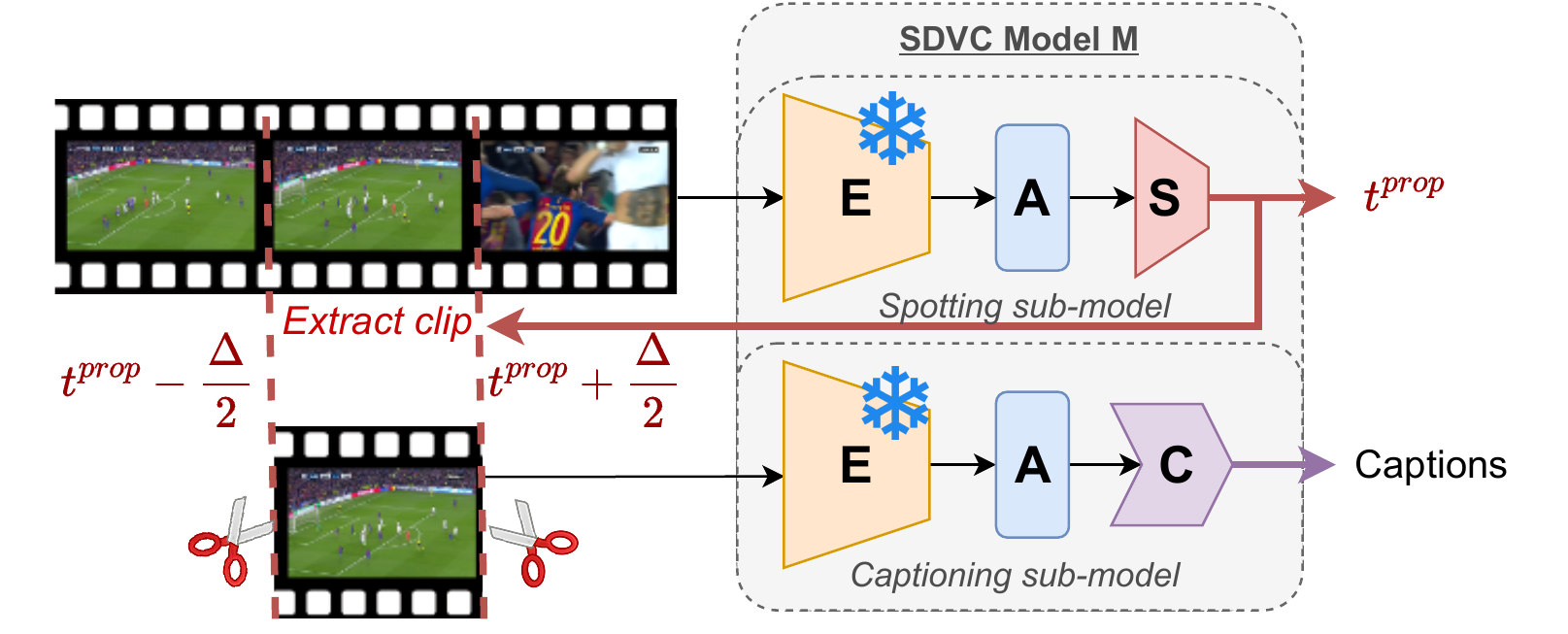}
    \caption{\textbf{Pipeline of our Single-anchored Dense Video Captioning (SDVC) baseline model $\mathbf{M}$}.  
    To generate dense comments with a single timestamp, we propose a two-stage approach. $\mathbf{M}$ consists of a spotting model followed by a captioning model. Both models use a shared frozen feature extractor $\mathbf{E}$ to generate a compact per-frame representation of the video. The spotting model uses an aggregator module $\mathbf{A}$ to combine the frame features into a single clip feature representation that is then passed to the spotting head $\mathbf{S}$ to generate proposal timestamps $t^{prop}$. The timestamps are then used to trim clips of size $\Delta$ that are subsequently processed by the captioning model through $\mathbf{E}$, $\mathbf{A}$, and a captioning head $\mathbf{C}$ to generate the anchored comment. 
    }
    \label{fig:pipeline}
\end{figure}

Following the literature on dense video captioning, we propose a two-stage approach~\cite{Zhou2018Towards, Krishna2017DenseCaptioning-arxiv, Iashin2020Multimodal-arxiv, Iashin2020ABetter-arxiv} as an initial baseline model for our SDVC task. 
Our model, denoted by $\mathbf{M}$, consists of a spotting model and a captioning model, which are cascaded together. The sub-models are trained independently and both consist of a frozen feature encoder $\mathbf{E}$, followed by an aggregator module $\mathbf{A}$, and either a spotting head $\mathbf{S}$ or a captioning head $\mathbf{C}$.
The feature encoder generates a compressed per-frame feature representation of the video clip, which is then temporally pooled by the aggregator. The resulting clip feature representation is subsequently passed either to the spotting head (locating where to generate the comments) or to the captioning head (generating a comment).
During inference, the spotting model generates a series of temporal proposals for a video:
$$\{t_0^{prop}, ..., t_N^{prop}\} = \mathbf{S}(\mathbf{A}(\mathbf{E}(\textrm{video})))\ ,$$
where $N$ is the total number of proposals.
The original video is then trimmed around each proposal $t_n^{prop}$ into a clip that is passed to the captioning model to generate a caption, following:
$$\textit{caption}_n = \mathbf{C}(\mathbf{A}(\mathbf{E}(\textrm{video}[t_n^{prop}-\frac{\Delta}{2}, t_n^{prop}+\frac{\Delta}{2}])))\ ,$$ where $\Delta$ is the window size of the captioning model.
Our baseline model is illustrated in Figure~\ref{fig:pipeline}.

\mysection{Feature encoder $\mathbf{E}$.}
We use the feature encoders provided in SoccerNet-v2, \ie ResNet-152~\cite{He2016Deep}, I3D~\cite{Carreira2017QuoVadis-arxiv}, C3D~\cite{Tran2015Learning}, and Baidu~\cite{Zhou2021Feature}.
The features are extracted at $1$ or $2$ fps from the original soccer broadcast videos.
%
To speed up the training, we reduce the feature dimensionality to $512$ for each image using PCA for the first three encoders, and a linear transformation for the Baidu features as suggested in~\cite{Giancola2021Temporally}.

\mysection{Aggregator $\mathbf{A}$.}
The aggregator module pools the frame feature vectors into one single compact feature representation of the clip.
For our baseline, we use four trainable pooling modules proposed by Giancola~\etal~\cite{Giancola2021Temporally}: NetVLAD, NetRVLAD, and the temporally aware pooling modules NetVLAD++ and NetRVLAD++.

\mysection{Spotting head $\mathbf{S}$.}
We build our spotting head as a dense layer with sigmoid activation that outputs $2$ classes: the presence of comment (foreground), and absence of comment (background). 
During training, the video is randomly cropped into video chunks, and a binary cross-entropy loss function is applied to the output.
During inference, we split the whole half-time video into overlapping clips and concatenate the predictions over time. We then use a Non-Maximum Suppression (NMS) algorithm to reduce redundant spots in a specific time window.

\mysection{Captioning head $\mathbf{C}$.}
Our captioning head is composed of two fully connected layers with ReLU activation and dropout, and a vanilla LSTM~\cite{Hochreiter1997Long} module with softmax activation. 
The fully connected layers project the output of the aggregator from the video hidden space to the language hidden space.
The projected features are then used to initialize the hidden state of the LSTM module that outputs the list of word confidence scores.
During training, we use the cross-entropy loss between the predicted and ground-truth words.
To stabilize the learning process, we use the teacher forcing method presented by Graves~\etal~\cite{Graves2012Supervised}.
As the soccer vocabulary is much more specific than generic language, we do not use pre-trained word embeddings, but learn the whole language during the training.
During inference, we sample words with a greedy approach, \ie the next word is the one with the highest confidence.

\subsection{Training parameters}

To train and evaluate our model, we use SoccerNet-Caption without the ``fun fact'' and ``attendance'' commentaries as they describe out-of-the-game content. 
During training, we use the Adam optimizer~\cite{Kingma2014Adam-arxiv} with PyTorch's default $\beta$ parameters. We reduce the learning rate by a factor of $10$ when the validation loss plateaus during $10$ consecutive epochs, with an initial value of $10^{-3}$, and a stopping criterion of $10^{-6}$.
The temporal pooling module is initialized with $64$ clusters and the dimension of the hidden vector of the LSTM is $512$,
the dropout value after the fully connected layers is set to $0.4$, and the word embedding dimension is set to $256$. Finally, the size of the vocabulary in SoccerNet-Caption is $1{,}769$ words.

\subsection{Results}

We now first separately study the performance of our spotting model and the captioning module, then  provide the performance of the whole pipeline for our new SDVC task.

\begin{table}[t]
\renewcommand*{\arraystretch}{0.90}\centering
    \begin{tabular}{l|l||c|c|c}
                        &               & \multicolumn{3}{c}{mAP@ (\%)}        \\ \cline{3-5}
        Encoder        & Aggregator       &      5 &     30 &     60 \\ \midrule
        RN\_PCA         & NetVLAD       &    6.4 &   39.1 &   38.0 \\ 
        RN\_PCA         & NetVLAD++     &    5.3 &   41.3 &   39.4 \\ 
        RN\_PCA         & NetRVLAD      &    7.0 &   39.5 &   37.8 \\ 
        RN\_PCA         & NetRVLAD++    &    5.4 &   41.5 &   39.4 \\ \hline
        I3D\_PCA        & NetVLAD       &    4.5 &   33.0 &   33.3 \\ 
        I3D\_PCA        & NetVLAD++     &    6.7 &   34.8 &   34.8 \\ 
        I3D\_PCA        & NetRVLAD      &    4.1 &   32.8 &   32.3 \\ 
        I3D\_PCA        & NetRVLAD++    &    5.9 &   34.5 &   34.1 \\ \hline
        C3D\_PCA        & NetVLAD       &    4.8 &   38.1 &   37.1 \\ 
        C3D\_PCA        & NetVLAD++     &    4.5 &   40.7 &   39.1 \\ 
        C3D\_PCA        & NetRVLAD      &    6.9 &   39.3 &   38.1 \\ 
        C3D\_PCA        & NetRVLAD++    &    3.9 &   39.2 &   38.2 \\ \hline
        Baidu           & NetVLAD       &\bf10.5 &   42.1 &   40.7 \\ 
        Baidu           & NetVLAD++     &    6.3 &\bf44.5 &\bf41.8 \\ 
        Baidu           & NetRVLAD      &    6.7 &   39.5 &   38.7 \\ 
        Baidu           & NetRVLAD++    &    3.6 &   44.0 &   41.2 \\
    \end{tabular}
    \caption{
    \textbf{Spotting results.}
    We train our spotting model to detect and localize comments and compare different combinations of encoder and pooling modules.
    }
    \label{tab:spotting}
\end{table}

\begin{table}[!ht]

\renewcommand*{\arraystretch}{0.85}    
\centering
    \begin{tabular}{c|c||c|c|c}
        WS & NMS & \multicolumn{3}{c}{mAP@ (\%)}  \\ \cline{3-5}
         (s) &   (s) &       5  &      30  &      60  \\ \midrule
        15     &  10      &\bf12.5 &   35.3 &   34.9  \\ 
        15     &  30      &    9.3 &\bf49.4 &\bf47.0  \\ 
        15     &  60      &    8.5 &   46.2 &   41.2  \\ \hline
        30     &  10      &    5.3 &   29.2 &   27.7  \\ 
        30     &  30      &    2.5 &   45.7 &   43.0  \\ 
        30     &  60      &    1.7 &   46.5 &   40.2  \\ \hline
        45     &  10      &    6.8 &   27.0 &   26.1  \\ 
        45     &  30      &    5.2 &   47.8 &   41.7  \\ 
        45     &  60      &    2.6 &   45.4 &   39.8  \\ \hline
        60     &  10      &    4.9 &   19.0 &   19.5  \\ 
        60     &  30      &    1.9 &   38.0 &   35.2  \\ 
        60     &  60      &    1.1 &   39.5 &   35.6  \\
    \end{tabular}
    \caption{
    \textbf{Spotting window size and NMS.}
    We compare several window and NMS sizes. Small windows achieve better results.}
    \label{tab:spotting ablation}
\end{table}

\mysection{Commentary spotting results.}
We first study the use of different combinations of feature encoders and aggregators, by fixing both the size of the input video chunk and the NMS window to $15$ seconds. 
The results are presented in Table~\ref{tab:spotting}.
As can be seen, the best results are obtained using the Baidu feature encoder and NetVLAD or NetVLAD++ pooling.
This suggests that using a feature extractor fine-tuned on soccer data leads to better performance even for commentary spotting.
However, there is not such a clear tendency regarding the aggregator module layer, with a slight advantage to NetVLAD or NetVLAD++.
Hence, for the following spotting model, we use the Baidu feature encoder with NetVLAD as it provides the best mAP@5.

Next, we study the influence of the window and NMS size on the quality of the localization.
As shown in Table~\ref{tab:spotting ablation}, the best performance is obtained with a window of $15$ seconds and a NMS of $10$ or $30$ seconds.
As the performance (mAP@30 and mAP@60) largely decreases for a NMS window of $10$ seconds, we choose $15$ and $30$ seconds for the window and NMS size of the following experiments. 
%
Our analysis shows that the commentary spotting performances are significantly lower than action spotting on SoccerNet-v2~\cite{Giancola2022SoccerNet}, as a comment may describe several actions, unlike spotting that focuses on a single action. 

\begin{table}[t]
\renewcommand*{\arraystretch}{0.90}
    \centering
     \resizebox{\linewidth}{!}{
    \begin{tabular}{l|l||c|c|c|c}
        Encoder  & Aggregator  &   B@4  &    M  &     R  &      C \\ \midrule
        RN\_PCA  & NetVLAD     &   4.0 &   21.8 &   21.5 &   10.8 \\ 
        RN\_PCA  & NetVLAD++   &   4.4 &   22.3 &   21.4 &   10.9 \\ 
        RN\_PCA  & NetRVLAD    &   4.2 &   21.9 &   21.4 &   10.6 \\ 
        RN\_PCA  & NetRVLAD++  &   4.2 &   21.8 &   21.2 &   11.0 \\ \hline
        I3D\_PCA & NetVLAD     &   3.2 &   21.9 &   20.4 &    7.5 \\ 
        I3D\_PCA & NetVLAD++   &   2.9 &   20.0 &   18.3 &    6.6 \\ 
        I3D\_PCA & NetRVLAD    &   3.1 &   21.1 &   20.7 &    8.0 \\ 
        I3D\_PCA & NetRVLAD++  &   3.7 &   21.7 &   20.8 &    9.2 \\ \hline
        C3D\_PCA & NetVLAD     &   3.9 &   21.9 &   21.4 &   10.3 \\ 
        C3D\_PCA & NetVLAD++   &   3.9 &   21.2 &   20.4 &   10.6 \\ 
        C3D\_PCA & NetRVLAD    &   4.0 &   21.6 &   21.1 &   11.0 \\ 
        C3D\_PCA & NetRVLAD++  &   3.9 &   21.3 &   20.7 &   11.0 \\ \hline
        Baidu    & NetVLAD     &   5.7 &\bf23.6 &\bf23.5 &   15.7 \\ 
        Baidu    & NetVLAD++   &   5.4 &   23.2 &   23.4 &\bf17.0 \\ 
        Baidu    & NetRVLAD    &   5.5 &   23.5 &   23.2 &   15.4 \\ 
        Baidu    & NetRVLAD++  &\bf5.7 &   23.5 &   23.3 &   16.0 \\    
    \end{tabular}
    }
    \caption{
    \textbf{Captioning results.}
    We train our captioning model to generate comments and compare different combinations of encoder and pooling modules with the Bleu (B), METEOR(M), ROUGE-L (R), and CIDEr (C) metrics.}
    \label{tab:ResultsCaptioning}
\end{table}

\begin{table}[t]
\renewcommand*{\arraystretch}{0.90} \centering
    \begin{tabular}{c||c|c|c|c}
        Window (s)  & B@4                & M            & R   & C    \\ \midrule
        15          & 5.4               & 23.5            & 23.2    & 15.2\\ \hline
        30          & 5.9               & \bf23.7   & 23.7    & 17.8\\ \hline
        45          & \bf5.9      & 23.6            & \bf{23.9}    & \bf{18.2} \\ \hline
        60          & 5.6               & 23.0           & 23.6   & 17.5 \\
    \end{tabular}
    \caption{
    \textbf{Captioning window size.}
    We compare several window sizes with the Bleu (B), METEOR (M), ROUGE-L (R), and CIDEr (C) metrics.}
    \label{tab:caption_time_ablation}
\end{table}

\begin{table}[t]
\renewcommand*{\arraystretch}{0.90} \centering
    \begin{tabular}{c|c||c|c|c|c}
        L   & TF ratio       & B@4              & M            & R       & C \\ \midrule
        2           & 0.5                 & 2.1             & 21.5            & 17.5                & 5.5 \\ 
        2           & 1                   & 6.0             & \bf23.7   & 23.8                & 17.8\\ \hline
        4           & 0.5                 & 4.1             & 22.4            & 21.7                & 8.8\\ 
        4           & 1                   & \bf6.0    & \bf23.7            & \bf24.1       & \bf18.5\\ \hline
        8           & 0.5                 & 1.3             & 18.1            & 15.8                & 3.4\\ 
        8           & 1                   & 2.5             & 20.7            & 21.6                & 7.3\\ 
    \end{tabular}
    \caption{
    \textbf{Captioning ablation.}
    We compare several numbers of LSTM layers (L) and teacher forcing (TF) ratio with the Bleu (B), METEOR (M), ROUGE-L (R), and CIDEr (C) metrics.  }
    \label{tab:ResultsCaptioning_teacher_layers}
\end{table}

\begin{table*}[ht!]

\renewcommand*{\arraystretch}{0.90}
    \centering
     \resizebox{\linewidth}{!}{
    \begin{tabular}{l||c|c|c||c|c|c|c||c|c|c|c||c}
& \multicolumn{3}{c||}{ {\bf Spotting} (mAP@ (\%))} & \multicolumn{4}{c||}{\bf Captioning} & \multicolumn{5}{c}{\bf Single-anchored dense video captioning}\\ \cline{2-13} 
Aggregator                                           &   5    &   30     &   60     &   B@4  &    M     &    R     &    C     &   B@4@30  &   M@30   &   R@30   &   C@30   & SODA\_c \\\hline
$\mathbf{A}_{                       }^{scratch}$     &\bf9.24 &   49.40  &\bf46.97  &   6.37 &\bf23.85  &   24.04  &   18.70  &   21.07   &   27.94  &   22.06  &   27.02  &   7.72 \\ \hline
$\mathbf{A}_{spot. \rightarrow capt.}^{frozen}$      &   2.99 &   49.06  &   46.32  &   6.04 &   23.68  &   24.05  &   18.53  &   21.10   &   27.24  &   21.99  &   26.96  &   7.71 \\ \hline
$\mathbf{A}_{spot. \rightarrow capt.}^{fine-tuned}$  &   4.67 &\bf49.61  &\bf46.97  &   6.18 &   23.71  &   24.03  &   19.00  &   21.03   &   28.14  &   22.03  &   27.01  &   7.76 \\ \hline
$\mathbf{A}_{capt. \rightarrow spot.}^{frozen}$      &   9.23 &   49.49  &\bf46.97  &   5.50 &   23.03  &   23.11  &   17.20  &   21.52   &   24.99  &   21.46  &   26.63  &   7.44 \\ \hline
$\mathbf{A}_{capt. \rightarrow spot.}^{fine-tuned}$  &\bf9.25 &   49.40  &\bf46.97  &\bf6.62 &   23.84  &\bf24.53  &\bf21.45  &\bf21.63   &\bf29.44  &\bf22.09  &\bf27.20  &\bf7.79 
    \end{tabular}
    }
    \caption{
    \textbf{SDVC Results.}
    We train SDVC models using Baidu encoder, NetVLAD as pooling layer, a window size of $15$ seconds for the spotting model, a NMS of $30$ seconds, a window size of $45$ seconds for the captioning module, four LSTM layers and with teacher forcing. We study different pre-trainings for the feature aggregators where the spotting and captioning models are trained separately from scratch, or the model is trained first on spotting (resp. captioning), and then weights are transferred to captioning (resp. spotting). The captioning (resp. spotting) aggregator is then either fine-tuned or frozen. 
    We report the spotting mAP @5/30/60, the captioning Bleu (B), METEOR (M), ROUGE-L (R), and CIDEr (C) with the corresponding single-anchored dense video captioning metrics @30, and the SODA metric.  
    }
    \label{tab:final results}
\end{table*}

\mysection{Captioning results.} 
We conduct similar ablations on the captioning task. 
As shown in Table~\ref{tab:ResultsCaptioning}, the best results are also achieved using the Baidu feature encoder and NetVLAD pooling.
This suggests again that pre-training a video encoder on the spotting task helps generating better commentaries.
In Table~\ref{tab:caption_time_ablation}, we also study the effect of the window size on the captioning model. Unlike the commentary spotting task, the best results are obtained with a window of $45$ seconds, showing that captioning requires more contextual information.
For the following experiments, we use $45$-seconds video clips as input to the captioning model.

As a final ablation study, we compare the performance of our captioning model when changing the number of stacked LSTM layers or the teacher forcing ratio.
The teacher-forcing ratio is the probability of the LSTM module receiving the true output sequence rather than its own prediction. 
For example, a teacher-forcing ratio of $0.5$ means that half of the time, the LSTM receives the true output sequence, and half of the time, it receives its own prediction.
As shown in Table~\ref{tab:ResultsCaptioning_teacher_layers}, the best performance is achieved with $4$ LSTM layers and using the teacher forcing technique.

\begin{figure}[t]
    \centering
    \includegraphics[width=0.85\linewidth]{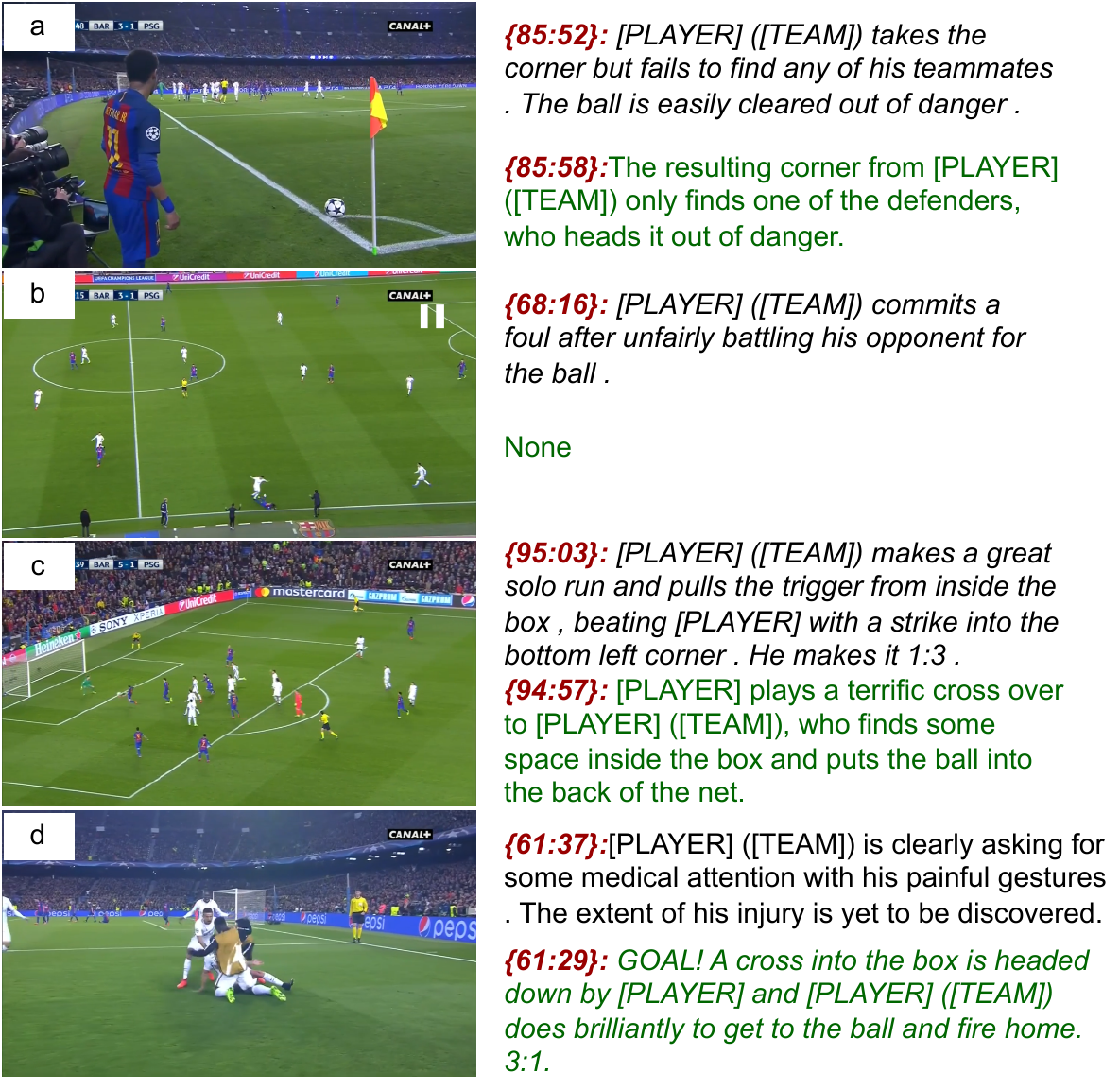}
    \caption{\textbf{Qualitative results.} (a) Our baseline is able to provide accurate captions for some actions with rich vocabulary, but (b) sometimes generates captions on unlabeled events.  (c) The scores are almost never correctly generated since the model generates captions based on clips without long temporal context and (d) sometimes generates completely wrong captions.}
    \label{fig:qualitative}
\end{figure}


\mysection{Single-anchored dense video captioning results.}
Finally, we evaluate the performance of our whole pipeline. 
Particularly, we study the influence of pre-training the aggregator in three ways: (1) no-pre-training for the spotting and captioning aggregator, (2) training the spotting model and transferring the weights to the captioning model, and (3) training the captioning model and transferring the weight to the spotting model.
We also either freeze or fine-tune the transferred weights on the second task.
As can be seen from Table~\ref{tab:final results}, the best performance for our SDVC task is obtained when training the captioning model from scratch, transferring the aggregator weights to the spotting aggregator, and fine-tuning those weights on the spotting task.



\mysection{Qualitative Results.}
We provide four predictions of our SDVC baseline in Figure~\ref{fig:qualitative} and compare them with the ground truth: (a)~Our method is able to generate good commentaries for some actions.
(b)~Our spotting model shows a tendency to generate proposals not close to any commentary, yet our captioning model still describes the ongoing action perfectly. This is also shown in the difference between the METEOR@30 and SODA\_c performances in Table~\ref{tab:final results}.
However, in a real-world application, such captions would add value rather than be considered a mistake. 
%
%
(c)~When generating a caption, our pipeline considers only the short time window around the proposal, hence
 the generated scores after a goal are almost never accurate.
(d)~We show a case of hard failure, where our model confuses a serious injury situation while the team was actually celebrating a goal.
These results show that our baseline is already able to generate accurate captions, but that there is room for improvement, especially in gathering temporal context.

\section{Conclusion}
\label{sec:conclusion}

This paper proposes the novel task of single-anchored dense video captioning focusing on generating textual commentaries anchored with single timestamps. To support this task, we present SoccerNet-Caption, a challenging dataset consisting of 37k timestamped commentaries across 715.9 hours of soccer broadcast videos. We benchmarked a first baseline algorithm on this dataset, highlighting the difficulty of temporally anchoring commentaries yet showing the capacity to generate meaningful commentaries. 

{

\mysection{Acknowledgement.}
This work was partly supported by KAUST OSR through the VCC funding and the SDAIA-KAUST Center of Excellence in Data Science and Artificial Intelligence. 
A. Cioppa is funded by the F.R.S.-FNRS.

}

{\small
\bibliographystyle{ieee_fullname}
\bibliography{bib/abbreviation-short, bib/bib, bib/dataset, bib/labo, bib/soccer, bib/sports}
}

\end{document}